\documentclass[letterpaper]{article}
\usepackage[preprint]{aaai2027}
\usepackage[hyphens]{url}
\usepackage{graphicx}
\usepackage{natbib}
\usepackage{caption}
\usepackage{algorithm}
\usepackage{algorithmic}
\usepackage{booktabs}

\title{When to Call an LLM: A Confidence-Gated Hybrid for Cost-Effective Emotion Recognition in Conversational AI}

\author{
    Sai Babu Udayagiri,
    Arjun Chouhan,
    Ravisekhar Kanagala,
    Trishala Pavagada
}
\affiliations{
    Five9\\
    saiudayagiri.babu@five9.com,\
    arjun.chouhan@five9.com,\
    Ravi.Kanagala@five9.com,\
    trish.pavagada@five9.com
}

\begin{document}

\maketitle

\begin{abstract}
Emotion recognition in conversation (ERC) is a production capability behind
agent-assist prompts, escalation routing, and post-call analytics in
contact-center-as-a-service (CCaaS) platforms, where cost and latency
constraints matter as much as accuracy. We report a systems-level
comparison of three deployment options for dialogue-contextual ERC: a
low-cost stacked ensemble (sentence embeddings, windowed context,
RandomForest/XGBoost/logistic-regression stacking), off-the-shelf LLM
prompting (GPT-4o-mini; zero-shot, few-shot, chain-of-thought), and a
confidence-gated hybrid that escalates only the ensemble's least-confident
predictions to the LLM -- a design directly modeled on IVA-to-human-agent
escalation policies already used in production contact centers. On
IEMOCAP, the ensemble significantly outperforms every LLM configuration
(0.595 vs.\ 0.460--0.536 weighted F1, $p<0.0001$) at a fraction of the cost
and sub-10ms latency; on MELD and CMU-MOSI the ranking reverses, showing
neither pure system is a universally safe deployment default. The
confidence-gated hybrid resolves this by Pareto-dominating both pure
systems on all three datasets (0.620, 0.643, 0.824 weighted F1) while
routing the majority of traffic through the near-zero-cost ensemble,
translating to roughly \$10--85 per million utterances versus
\$99--170 for an LLM-only pipeline. The escalation policy is not an opaque
cost/accuracy dial: escalated turns disproportionately follow an emotion
or sentiment shift, giving operators an interpretable, auditable routing
signal, and the ensemble's confidence is independently well-calibrated and
safely under- rather than over-confident. The pattern holds across three
datasets and two independent LLM providers. Confidence-gated cascading is
an established idea in general ML systems; our contribution is showing it
transfers cleanly to dialogue-contextual ERC and yields a concrete,
actionable deployment recipe for CCaaS and conversational-AI platforms
deciding how to allocate LLM spend.
\end{abstract}

\section{Introduction}

Emotion recognition in conversation (ERC) underlies real-time agent-assist
prompts, automated escalation routing, quality-assurance scoring, and
post-call analytics in CCaaS platforms. Production systems rarely have the
latency and throughput budget for large sequence/graph ERC models, which
motivated our prior work \citep{parida2023vartarasa} showing a comparatively
simple stacked ensemble of tree-based learners over sentence embeddings
and windowed dialogue context can match more complex architectures on
IEMOCAP at a fraction of the cost. Since then, general-purpose LLMs have
become the default ``try first'' tool for text classification, typically
via zero-shot or few-shot prompting. This raises a deployment-relevant
question: does off-the-shelf LLM prompting actually outperform a
well-engineered lightweight ensemble on dialogue-contextual ERC, and if
the answer is dataset-dependent, what deployment architecture avoids
betting the wrong way?

The deployment options above are not the only two: state-of-the-art
dialogue-contextual ERC architectures -- EmoBERTa \citep{kim2021emoberta},
Joyful \citep{li2023joyful}, MiSTER-E \citep{dutta2026mistere} -- use
fine-tuned transformer backbones or multimodal (text+audio+visual)
fusion, incurring a per-domain fine-tuning cost and GPU-serving
latencies that routinely exceed real-time p95 SLA budgets for
agent-assist (\S2). Our ensemble (\S3) requires neither fine-tuning nor
GPU serving and runs in sub-10ms on CPU; an LLM API call incurs neither
a fine-tuning cost nor a GPU-serving requirement but is priced and
latency-bound per call. The deployment choice is therefore three-way,
not two-way, and this paper brackets the fine-tuned/multimodal option:
its accuracy ceiling and cost profile are established in the existing
ERC literature (\S2, Appendix~\ref{app:sota}), leaving open only whether,
between the two options with no GPU-serving cost, an LLM call is
necessary or a calibrated lightweight ensemble already suffices.

Confidence-gated cascading -- a cheap model handling most traffic, an
expensive one called only on hard cases -- is an established pattern in
general ML systems \citep{chen2023frugalgpt,nie2024online}. Our
contribution is not the cascading idea itself, but showing it transfers
cleanly to dialogue-contextual ERC when the cheap stage is a calibrated,
non-LLM ensemble rather than a smaller LLM or a learned deferral policy,
and characterizing exactly what such a router learns to escalate.
Concretely, we:

\begin{enumerate}
\item Provide a controlled, dialogue-level comparison of a lightweight
  ensemble against GPT-4o-mini across three prompting strategies on
  IEMOCAP, MELD, and CMU-MOSI, with paired bootstrap significance testing,
  showing neither pure system is a safe default across datasets.
\item Identify and empirically confirm a specific mechanism -- chain-of-
  thought prompting's systematic under-prediction of the majority
  ``neutral'' label -- explaining why CoT is IEMOCAP's best LLM strategy
  but MELD's worst (Appendix~\ref{app:cot-neutral}).
\item Propose and validate a confidence-gated hybrid that routes on the
  ensemble's own calibrated class probability, Pareto-dominating both
  pure systems on all three datasets and both of two independent LLM
  providers tested.
\item Show the router is interpretable, not a black-box threshold:
  escalated turns disproportionately follow an emotion/sentiment shift (a
  modest but consistent 1.11--1.17$\times$ lift), and the underlying
  ensemble confidence is independently well-calibrated and mildly
  under-confident -- the safety-favorable miscalibration direction for a
  production escalation gate.
\end{enumerate}

\section{Related Work}

\textbf{ERC architectures.} DialogueRNN \citep{majumder2019dialoguernn},
DialogueGCN \citep{ghosal2019dialoguegcn}, and COSMIC
\citep{ghosal2020cosmic} model conversational context via recurrence,
graph structure, and commonsense knowledge respectively. More recent work
moves toward large fine-tuned transformers and multimodal/LLM-embedding
fusion: EmoBERTa \citep{kim2021emoberta}, Joyful \citep{li2023joyful}, and
MiSTER-E \citep{dutta2026mistere}. These benchmark against each other on
the premise that architectural sophistication is necessary; lightweight
tree-ensembles have been shown competitive with, though not superior to,
this lineage at a fraction of the cost \citep{parida2023vartarasa}, but no
prior published ERC work compares such an ensemble against LLM prompting
or a confidence-gated hybrid of the two. This is a practically important
gap for CCaaS platforms specifically: recurrence-, graph-, and
LLM-embedding-fusion architectures generally require GPU serving and
per-utterance latencies well outside typical p95 SLA budgets for
real-time agent-assist, so a system's benchmark accuracy is only one of
several deployment-relevant properties, alongside training cost,
inference latency, and interpretability of its predictions.

\textbf{LLMs for emotion/sentiment classification.} Most LLM-prompting
benchmarks for affect classification evaluate single-utterance sentiment,
not dialogue-contextual ERC. The rare dialogue-level exception,
\citet{feng2024affect}, evaluates zero-shot/few-shot LLM prompting against
fine-tuned neural ERC baselines and finds LLMs generally underperform --
the closest existing empirical result to our own finding that off-the-shelf
LLM prompting does not uniformly beat a non-LLM ERC baseline (\S4), obtained
with a different (fine-tuned, not tree-ensemble) baseline type. LLM
prompting also degrades substantially in multi-turn settings generally
\citep{laban2025lost}, and the value of chain-of-thought for affect tasks
specifically is contested: it helps in some sentiment-analysis settings
and not others \citep{zheng2025reassessing,miriyala2025enhancing},
mirroring our own finding that CoT is IEMOCAP's best strategy but MELD's
worst.

\textbf{Cost-aware cascading and confidence-based routing.} This
literature splits into three lines. \emph{LLM-to-LLM cascades and learned
routers} evaluate or generate a candidate answer first, then decide
whether to escalate: FrugalGPT \citep{chen2023frugalgpt} gates on a
learned answer-quality scorer; RouteLLM \citep{ong2024routellm} learns a
win-probability model from preference data rather than a calibrated
score; AutoMix \citep{aggarwal2023automix} pairs few-shot
self-verification with a POMDP router; Tryage \citep{sikka2023tryage}
predicts downstream performance per prompt; EcoAssistant
\citep{zhang2023ecoassistant} escalates on code-execution failure rather
than a probability signal; Mixture-of-Agents \citep{wang2024moa} layers
multiple LLMs with no escalation decision at all, a useful contrast case.
\citet{dekoninck2024unified} formalize cascade routing theoretically and
show a quality estimator's own accuracy is the binding constraint on
cascade performance -- direct motivation for gating on a well-calibrated
non-LLM estimator, as we do, rather than an LLM-judged one. A second line
escalates a \emph{non-LLM} model to an LLM via a confidence-style signal,
architecturally closer to our design: \citet{nie2024online} defer from
logistic regression to an LLM via a learned, continuously-updated
deferral policy for general streaming inference and discuss simple
confidence-threshold deferral only as a baseline; Gatekeeper
\citep{rabanser2025gatekeeper} adds a calibration-tuning loss so a
cascade's cheap model is confident exactly when correct, direct support
for why calibration quality (which we verify empirically,
Appendix~\ref{app:calibration}) matters for this design; Confidence
Tokens \citep{chuang2024confidencetokens} instead trains the LLM itself
to emit a routing token, sitting between the two lines above;
\citet{fanconi2025cascaded} extend escalation to a three-tier base
model/LLM/human cascade for decision-making. A third, theoretical line
grounds when threshold-based deferral is justified in the first place:
\citet{fisch2022calibrated} formalize selective calibration error, and
\citet{hendrickx2021reject} survey reject-option classification broadly.
Our contribution relative to all three lines is task-specific and
architectural: gating directly on a calibrated tree-ensemble's own class
probability, with no learned deferral policy and no LLM-judged quality
score, transfers cleanly to dialogue-contextual ERC, Pareto-dominates
both pure endpoints across three datasets and two LLM providers, and
yields an interpretable escalation signal -- a modest but consistent
correlation with emotion/sentiment shifts (\S 4.4) -- with a direct cost
translation for CCaaS deployment decisions -- a combination not
previously demonstrated for ERC.

\section{Method}

\begin{table}[t]
\centering
\footnotesize
\begin{tabular}{@{}lcccc@{}}
\toprule
\textbf{Dataset} & \textbf{Classes} & \textbf{Train} & \textbf{Val} & \textbf{Test} \\
\midrule
IEMOCAP  & 6 & 5{,}081 & 677$^\dagger$ & 1{,}622 \\
MELD     & 7 & 9{,}989 & 1{,}109$^\ddagger$ & 2{,}610 \\
CMU-MOSI & 3 & 1{,}284 & 229$^\ddagger$ & 686 \\
\bottomrule
\end{tabular}
\caption{Dataset statistics. IEMOCAP \citep{busso2008iemocap} holds out
Session 5 as test, matching standard convention; MELD
\citep{poria2019meld} and CMU-MOSI \citep{zadeh2016mosi} use their
released train/test splits. $^\dagger$IEMOCAP's Val is not an official
split -- we carve it from the training dialogues ourselves (10\%) for
development-time sanity checks. $^\ddagger$MELD's and CMU-MOSI's Val
columns are each dataset's own released dev/valid split; none of the
three datasets' Val rows are merged into Train or used in any reported
result -- they are held out for diagnostic checks only, consistent with
standard practice for these benchmarks.}
\label{tab:datasets}
\end{table}

\textbf{Ensemble baseline.} Each utterance is encoded with a pretrained
sentence embedding (\texttt{all-MiniLM-L6-v2}, 384-dim, no fine-tuning),
concatenated with the mean embedding of the preceding 5 utterances and a
same-speaker/different-speaker flag, yielding a 769-dim feature vector.
The stacked ensemble's Level-0 base learners are two RandomForest
configurations (300 trees/unlimited depth/balanced class weighting; 500
trees/depth 20/balanced-subsample weighting) and two XGBoost
configurations (300 trees/depth 6/learning rate 0.1/0.8 subsample; 500
trees/depth 4/learning rate 0.05/0.9 subsample), originally selected for
ensemble diversity. We tune this choice directly in
Appendix~\ref{app:hpo} via a systematic hyperparameter search: a 20-trial
search on each dataset improves test-set weighted F1 by only
0.17--0.78 points, confirming these values were already close to
optimal rather than an arbitrary choice. A logistic-regression Level-1
meta-learner is trained on out-of-fold Level-0 class-probability
predictions via 5-fold internal cross-validation (scikit-learn
\texttt{StackingClassifier}, \texttt{stack\_method=}\texttt{"predict\_proba"}).
The ensemble is trained once per dataset on the full training split
(Table~\ref{tab:datasets}) and evaluated once on the full held-out test
split; it requires no GPU and runs in sub-10ms per utterance on CPU.

\textbf{LLM prompting.} We evaluate GPT-4o-mini under zero-shot, few-shot
(6-shot), and chain-of-thought (CoT, 6-shot) prompting, with a 5-turn
context window matching the ensemble, temperature 0, and logged
per-call cost, latency, and token counts. Appendix~\ref{app:cmumosi}
additionally evaluates Llama-3-8B-Instruct, on the full test split of
all three datasets, as an independent second provider.

\textbf{Confidence-gated hybrid.} Algorithm~\ref{alg:routing} formalizes
the routing rule: let $p_{\max}$ be the ensemble's top-class probability
and $\tau \in [0,1]$ a threshold. We sweep $\tau$ over $\{0.00, 0.05,
\ldots, 1.00\}$ to trace an accuracy-vs-cost curve against the two pure
endpoints ($\tau{=}0$: pure ensemble; $\tau{=}1$: pure LLM), and report
results at each dataset's empirically-best $\tau^*$. This sweep requires
a fully labeled evaluation split, which will not exist for a genuinely
new production dataset; Appendix~\ref{app:coldstart} gives a practical
cold-start procedure and simulates it directly on all three datasets.

\begin{figure*}[t]
\centering
\includegraphics[width=0.85\textwidth]{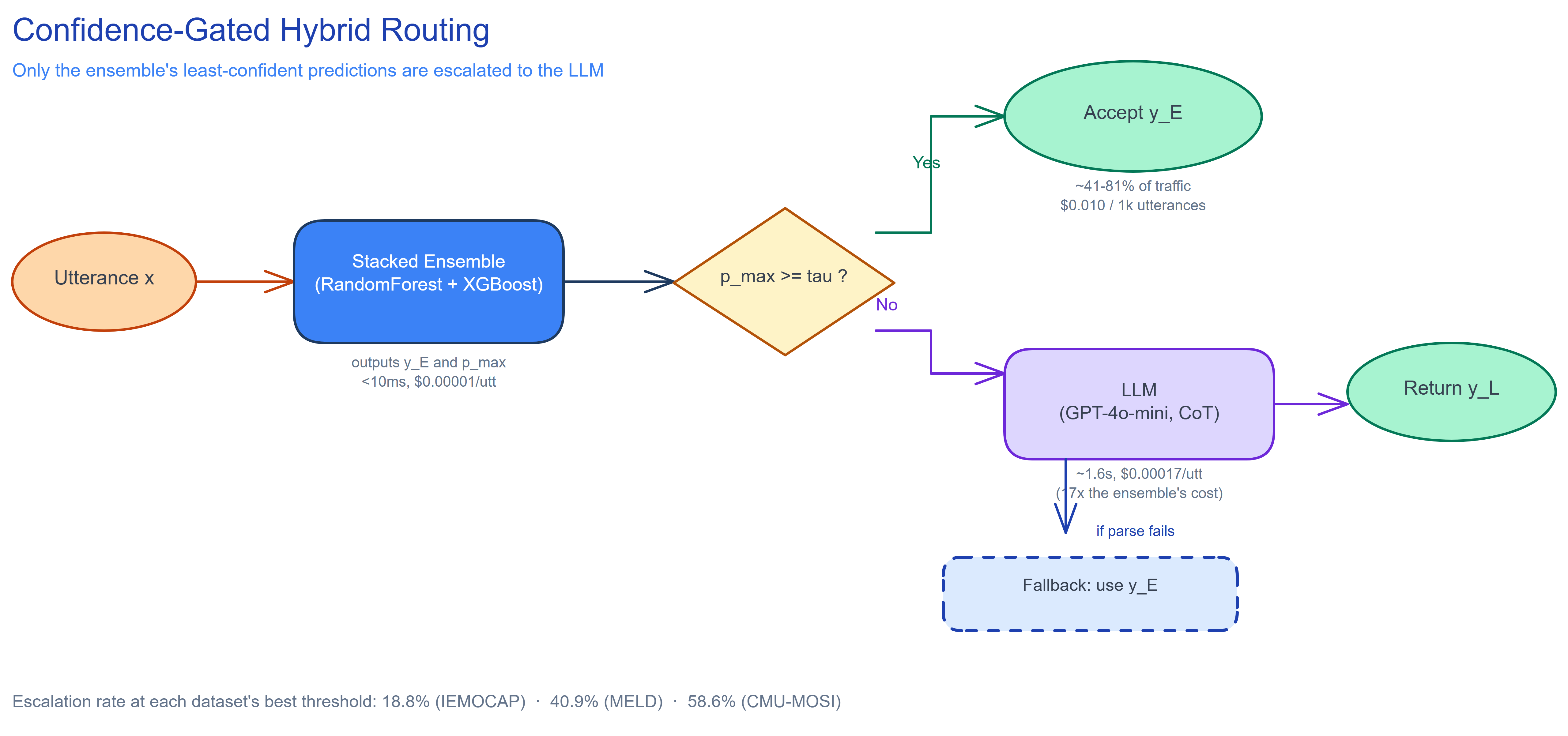}
\caption{Confidence-gated hybrid routing architecture. The stacked
ensemble classifies every utterance and outputs its own top-class
probability $p_{\max}$; only utterances below threshold $\tau$ are
escalated to the LLM. Evidence numbers are IEMOCAP, CoT strategy
(Table~\ref{tab:cost}); escalation rates are each dataset's own
$\tau^*$ (\S 4.4).}
\label{fig:architecture}
\end{figure*}

\begin{algorithm}[t]
\caption{Confidence-Gated Hybrid Routing}
\label{alg:routing}
\begin{algorithmic}[1]
\REQUIRE utterance $x$, ensemble $E$, LLM $L$, threshold $\tau$
\STATE $(\hat{y}_E, p_{\max}) \leftarrow E(x)$ \COMMENT{ensemble prediction + top-class prob.}
\IF{$p_{\max} \geq \tau$}
  \STATE \textbf{return} $\hat{y}_E$ \COMMENT{near-zero marginal cost}
\ELSE
  \STATE $\hat{y}_L \leftarrow L(x)$ \COMMENT{escalate; costs one LLM call}
  \IF{$\hat{y}_L$ parses successfully}
    \STATE \textbf{return} $\hat{y}_L$
  \ELSE
    \STATE \textbf{return} $\hat{y}_E$ \COMMENT{fallback on parse failure}
  \ENDIF
\ENDIF
\end{algorithmic}
\end{algorithm}

\textbf{Evaluation.} We report weighted F1 (primary, since all three
datasets are class-imbalanced to varying degrees) and macro F1
(secondary, equal weight per class), with paired bootstrap significance
testing between model pairs (1,000 resamples, seeded, two-sided
$p$-value and 95\% CI on the weighted-F1 difference), and three
failure-mode groupings computed from dialogue structure alone (no manual
annotation): context length, emotion-shift status, and minority-class
membership. We use the full IEMOCAP, MELD, and CMU-MOSI test splits
(Table~\ref{tab:datasets}) for every configuration -- no subsampling.

\section{Experiments and Results}

\subsection{Accuracy: Neither Pure System Is a Safe Default}

\begin{table}[t]
\centering
\scriptsize
\begin{tabular}{@{}lcccc@{}}
\toprule
\textbf{Dataset (strategy)} & \textbf{Ens.} & \textbf{LLM} & \textbf{$\Delta$ 95\%CI} & \textbf{$p$} \\
\midrule
IEMOCAP (ZS) & \textbf{.595} & .460 & [.10,.17] & $<$.0001 \\
IEMOCAP (CoT) & \textbf{.595} & .536 & [.03,.09] & $<$.0001 \\
MELD (FS)    & .561 & \textbf{.631} & [$-$.09,$-$.05] & $<$.0001 \\
CMU-MOSI (CoT) & .725 & \textbf{.814} & [$-$.13,$-$.05] & $<$.0001 \\
\bottomrule
\end{tabular}
\caption{Weighted F1, ensemble vs.\ best LLM strategy, full test splits,
with 95\% CI on the weighted-F1 difference $\Delta$ (Ens.\ $-$ LLM) from
paired bootstrap. The winning pure system flips by dataset.}
\label{tab:rq1}
\end{table}

On IEMOCAP the ensemble significantly beats GPT-4o-mini under every
prompting strategy (Table~\ref{tab:rq1}), with the LLM's disadvantage
concentrated in long-context turns and CoT narrowing but not closing the
gap. On MELD, zero-shot and few-shot prompting significantly \emph{exceed}
the ensemble -- driven by MELD's severe class imbalance, where the
ensemble collapses on minority classes (disgust/fear F1 = 0.033 vs.\ the
LLM's 0.533 on the same utterances) -- while CoT falls significantly
\emph{below} the ensemble. We trace this specific reversal to a
confirmed mechanism, not a MELD-specific quirk: CoT systematically
under-predicts the majority ``neutral'' label relative to few-shot on
\emph{both} IEMOCAP and MELD (Appendix~\ref{app:cot-neutral}), but this
bias only becomes fatal on MELD, where neutral is $\sim$48\% of the test
set versus IEMOCAP's $\sim$24\%. Critically, MELD's zero-/few-shot
reversal turns out to be \textbf{specific to GPT-4o-mini}: an
independently-tested Llama-3-8B loses to the ensemble on \emph{every}
MELD strategy, including zero-shot and few-shot
(Appendix~\ref{app:cmumosi}) -- so ``the LLM wins on MELD'' describes one
commercial model's behavior, not a provider-general property of the
dataset. On CMU-MOSI, by contrast, the LLM wins at every strategy against
both GPT-4o-mini and Llama-3-8B, and on IEMOCAP the ensemble wins against
both -- these two rankings \emph{are} provider-robust.
\textbf{The practical takeaway is that a CCaaS operator cannot safely
default to either pure system across tasks, and for at least one dataset
(MELD) not even across LLM vendors} -- exactly the deployment risk the
hybrid below is designed to eliminate.

\subsection{Where the Ensemble and LLM Disagree}

\begin{table}[t]
\centering
\footnotesize
\begin{tabular}{@{}lccc@{}}
\toprule
\textbf{IEMOCAP slice} & \textbf{$n$} & \textbf{Ens.} & \textbf{LLM (CoT)} \\
\midrule
Short context ($<$5)   & 155   & 0.537 & 0.633 \\
Long context ($\geq$15) & 1{,}157 & 0.598 & 0.518 \\
Stable (no shift)      & 941   & 0.687 & 0.583 \\
Emotion shift          & 681   & 0.464 & 0.473 \\
\bottomrule
\end{tabular}
\caption{IEMOCAP weighted F1 by context length and emotion-shift status.
The LLM wins short-context turns; the ensemble wins long-context turns.
Both degrade substantially on shift turns.}
\label{tab:failure-modes}
\end{table}

The two systems fail on different slices of the data
(Table~\ref{tab:failure-modes}), which is exactly what makes routing
between them useful rather than redundant: the LLM is more accurate on
short-context turns but trails on long-context dialogue -- consistent
with LLMs' documented tendency to lose track of earlier context in
multi-turn settings \citep{laban2025lost} -- while both systems degrade
substantially on emotion-shift turns, more so for the ensemble in
relative terms, consistent with the classical motivation for explicit
emotional-inertia modeling in ERC architectures. On MELD, the ensemble's
minority-class collapse (disgust/fear F1 = 0.033 across 118 test
utterances) accounts for most of its aggregate deficit against the
LLM (F1 = 0.533 on the same utterances) -- a structural weakness of a
frequency-driven tree ensemble under severe class imbalance that the
LLM's world-knowledge priors do not share. On CMU-MOSI, the disagreement
traces to a related but more extreme version of the same mechanism: the
ensemble's top predicted class is \emph{never} ``neutral'' on any test
utterance (Appendix~\ref{app:perclass}), so every genuinely neutral-
sentiment turn is forced into a polarized positive/negative call, while
the LLM's world-knowledge priors recognize understated or mixed
sentiment the frequency-driven ensemble structurally cannot select --
not merely a rare prediction, as with MELD's minority classes, but one
the ensemble's argmax never makes.

\subsection{Cost and Latency in Deployment Terms}

\begin{table}[t]
\centering
\footnotesize
\begin{tabular}{@{}lccc@{}}
\toprule
\textbf{Tier} & \textbf{IEMOCAP} & \textbf{MELD} & \textbf{CMU-MOSI} \\
\midrule
Ensemble & \$10/1M & \$10/1M & \$10/1M \\
Hybrid   & \$82/1M & \$61/1M & \$85/1M \\
Pure LLM & \$170/1M & \$99/1M & \$138/1M \\
\bottomrule
\end{tabular}
\caption{Cost per million utterances per deployment tier (ensemble cost
fixed at \$10/1M).}
\label{tab:cost}
\end{table}

\begin{figure*}[t]
\centering
\includegraphics[width=0.98\textwidth]{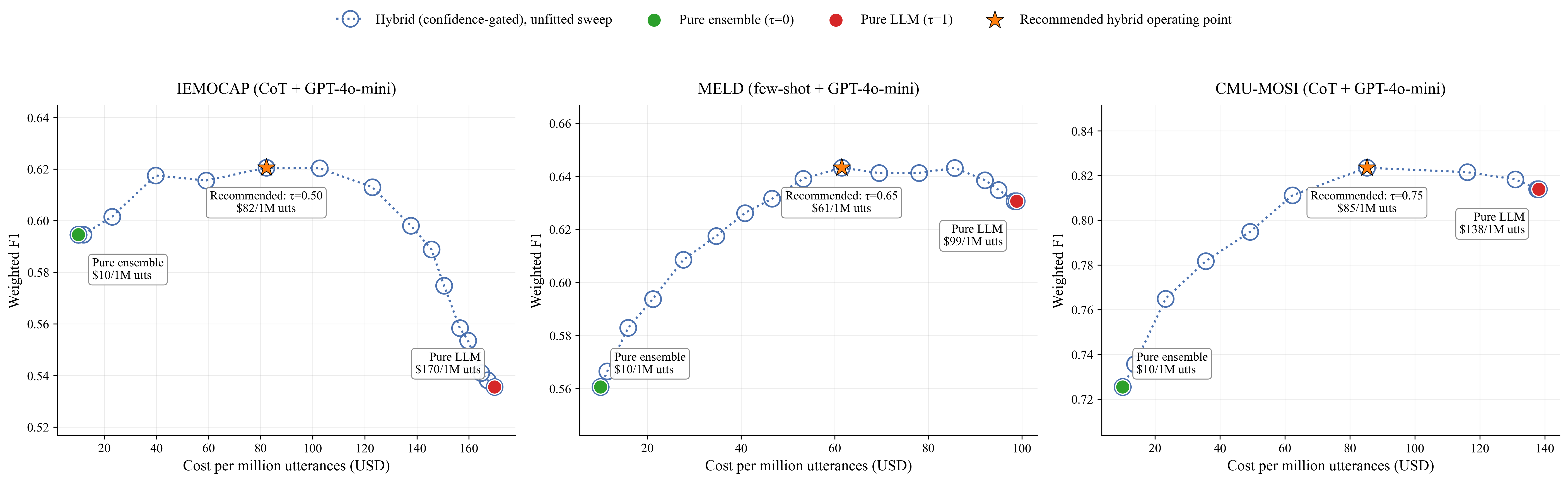}
\caption{Weighted F1 vs.\ cost per million utterances for IEMOCAP, MELD,
and CMU-MOSI, full test splits. Markers show the actual measured sweep
(dotted, unfitted line); the recommended hybrid operating point on each
dataset simultaneously beats the pure-ensemble and pure-LLM endpoints on
accuracy while costing a fraction of the pure-LLM tier.}
\label{fig:cost-accuracy}
\end{figure*}

Ensemble inference is CPU-bound and sub-10ms; every LLM call costs
$3.9$--$17\times$ the ensemble's per-utterance price and adds
sub-second-to-low-single-digit-second network latency, a direct tension
with real-time p95-latency SLAs in live agent-assist use cases. Re-cast in
deployment terms (Table~\ref{tab:cost}), a representative volume of 1M
utterances/month costs roughly \$10/month on the pure ensemble,
\$61--85/month on the recommended hybrid, and \$99--170/month on a
pure-LLM pipeline -- figures directly usable in a procurement
conversation, not just an abstract accuracy metric.

\subsection{The Confidence-Gated Hybrid Pareto-Dominates Both Endpoints}

\begin{table}[t]
\centering
\scriptsize
\begin{tabular}{@{}lccccc@{}}
\toprule
\textbf{Dataset} & \textbf{Ens.} & \textbf{Hybrid} & \textbf{Best LLM} & \textbf{$\Delta$ 95\%CI} & \textbf{$p$} \\
\midrule
IEMOCAP  & 0.595 & \textbf{0.620} & 0.536 & [.01,.05] & .02 \\
MELD     & 0.561 & \textbf{0.643} & 0.631 & [.00,.02] & .01 \\
CMU-MOSI & 0.725 & \textbf{0.824} & 0.814 & [$-$.01,.03] & .38 \\
\bottomrule
\end{tabular}
\caption{Weighted F1: pure ensemble vs.\ confidence-gated hybrid (at each
dataset's best threshold/strategy) vs.\ the better pure LLM. $\Delta$ is
hybrid minus the better of the two pure systems, paired bootstrap (1,000
resamples). The hybrid numerically wins on every dataset regardless of
which pure system otherwise wins, but the margin is only statistically
significant on two of three.}
\label{tab:rq4}
\end{table}

This is the paper's central result (Table~\ref{tab:rq4}, full sweeps in
Appendix~\ref{app:lowconf}): on \emph{every} dataset tested, confidence-gated
routing numerically beats \emph{both} pure endpoints simultaneously, not
merely trading accuracy for cost along a frontier. This holds even on
MELD and CMU-MOSI, where the pure ensemble loses the head-to-head
comparison -- the hybrid does not simply default to ``always call the
stronger pure system''; it extracts a genuine complementarity benefit
because the ensemble's confidence signal reliably identifies a subset of
utterances it handles correctly and cheaply, regardless of which system
wins in aggregate. We hold this claim to the same statistical standard as
Table~\ref{tab:rq1} rather than reporting only point estimates: the
hybrid's margin over the better pure system is significant on IEMOCAP
($p=.02$) and MELD ($p=.01$), but \textbf{not} on CMU-MOSI ($p=.38$,
95\% CI crosses zero) -- our smallest test split (686 utterances) is
underpowered to statistically distinguish a 1-point accuracy margin. On
CMU-MOSI specifically, the honest claim is narrower than ``more
accurate'': the hybrid is statistically indistinguishable in accuracy
from the pure LLM while costing a fraction as much (\$85 vs.\ \$138 per
million utterances, Table~\ref{tab:cost}) -- a cost-dominance claim,
which is unambiguous, standing in for an accuracy-dominance claim we
cannot fully support at this sample size. The optimal escalation rate
varies by dataset (45.4\% IEMOCAP, 57.5\% MELD, 58.6\% CMU-MOSI) but the
qualitative pattern -- a broad plateau of $\tau$ values beating both
endpoints -- is robust across all three datasets and, on CMU-MOSI
specifically, across both LLM providers tested
(Appendix~\ref{app:cmumosi}).

\subsection{The Router Is Interpretable, Not a Black Box}

\begin{table}[t]
\centering
\footnotesize
\begin{tabular}{@{}lccc@{}}
\toprule
\textbf{Dataset} & \textbf{$\tau^*$} & \textbf{No shift} & \textbf{Shift} \\
\midrule
IEMOCAP  & 0.50 & 39.7\% (0.87$\times$) & 53.3\% (1.17$\times$) \\
MELD     & 0.65 & 50.8\% (0.88$\times$) & 63.8\% (1.11$\times$) \\
CMU-MOSI & 0.75 & 54.0\% (0.92$\times$) & 66.3\% (1.13$\times$) \\
\bottomrule
\end{tabular}
\caption{Escalation rate by emotion/sentiment-shift status, at each
dataset's $\tau^*$. Shift turns escalate at a consistent 1.11--1.17$\times$
lift over the dataset's overall rate.}
\label{tab:shift-lift}
\end{table}

Escalated turns disproportionately follow an emotion or sentiment shift
from the preceding turn (Table~\ref{tab:shift-lift}), with a consistent,
if modest, 1.11--1.17$\times$ lift across all three datasets regardless
of task granularity or LLM provider -- of four turn-level properties
tested (shift status, utterance length, minority-class membership,
speaker-initial position; full breakdown in Appendix~\ref{app:lowconf}),
only shift status is directionally consistent across all three. We are
deliberately conservative in how we describe this: a 1.11--1.17$\times$
lift means escalation is somewhat, not overwhelmingly, more likely on
shift turns, and we do not claim the router explicitly represents or
detects emotional dynamics -- only that its confidence signal, with no
emotional-dynamics modeling built into the architecture, happens to
correlate with them, directly corroborating the failure-mode finding
above that both systems, and the ensemble especially, struggle on shift
turns. This is independently
corroborated by a calibration analysis (Appendix~\ref{app:calibration}):
the ensemble's confidence is well-calibrated on all three datasets
(ECE $\leq 0.046$) and mildly \emph{under}-confident, the safety-favorable
miscalibration direction for a production escalation gate -- it
occasionally sends an already-correct prediction to the LLM (a small
cost inefficiency) rather than silently keeping a wrong prediction the
ensemble is falsely confident about.

\section{Discussion}

For CCaaS platforms deciding how to allocate LLM spend on emotion-related
classification, three findings are directly actionable. First, the
IEMOCAP-vs-MELD/CMU-MOSI reversal means a lightweight ensemble is not a
universally safe first choice, and neither is off-the-shelf LLM prompting
-- the deployment decision is task-specific, not a blanket rule. Second,
the confidence-gated hybrid removes the need to make that bet: it wins on
every dataset we tested by escalating only the fraction of traffic the
cheap system is unsure about, at a fraction of full-LLM cost. Third, the
escalation signal is auditable in operational terms (it correlates with
emotion/sentiment shifts, a concept CCaaS quality teams already reason
about for escalation-worthy calls) rather than an opaque probability
cutoff, which matters for teams that need to explain and tune automated
routing decisions to stakeholders. We view this as a concrete instance of
a more general principle for hybrid system design: look for
structurally distinct failure modes between a candidate cheap and
expensive system, since the size of the benefit here is a direct
consequence of the two systems disagreeing on \emph{which} utterances are
hard, not simply one being uniformly better.

This also points to a concrete direction for improving the LLM side of
the pipeline rather than only the routing policy: the LLM's long-context
weakness on IEMOCAP is consistent with LLMs' broader tendency to lose
track of earlier content in multi-turn settings \citep{laban2025lost}, so
explicit conversational-state tracking injected into the prompt (e.g., a
running emotion-trajectory summary, analogous to recurrence-based ERC
architectures' speaker-state mechanism) may narrow the gap the hybrid
currently has to route around, rather than extending the raw context
window or relying on chain-of-thought reasoning depth, which our results
show is an inconsistent lever for affect classification specifically
\citep{zheng2025reassessing,miriyala2025enhancing}. For teams already
operating an escalation-based CX platform, the practical entry point is
low-risk: the routing policy requires no LLM fine-tuning and no change to
existing IVA-to-human handoff infrastructure, only a calibrated
confidence score from whatever classifier already sits in front of the
LLM call today.

The main limitation is generality of the cheap-model type: we validate a
tree-ensemble-to-LLM cascade specifically for ERC; whether the same
recipe (calibrated non-LLM confidence, no learned deferral policy) holds
for other dialogue-structured classification tasks is untested. We do,
however, test a narrower but directly relevant version of this question --
does the benefit survive a \emph{stronger} cheap classifier on the
\emph{same} task -- in Appendix~\ref{app:strongerbase}, substituting an
independently-developed fine-tuned transformer for our ensemble: the
answer is mixed rather than uniformly positive, reinforcing that this
paper's routing benefit should be read as contingent on a documented,
classifier-specific failure mode rather than as a property of gated
routing for ERC in general. A fuller discussion of limitations, baseline
fidelity, and open questions (e.g., MELD's CoT reversal) is in
Appendix~\ref{app:limitations}.

The most important open question for CCaaS deployment specifically is
whether these results transfer from academic benchmarks (acted dialogue,
scripted TV drama, unscripted vlogs) to real contact-center transcripts,
which we have not piloted on. We cannot close this gap with new data
here, but the transfer risk is more bounded than it first appears once
placed against the literature: ASR-transcript noise at the word-error
rates realistic ASR systems produce on emotional speech leaves accuracy
on IEMOCAP and CMU-MOSI -- the two datasets in our own evaluation --
statistically on par with clean human transcripts
\citep{li2024serasr}; LLMs are the more out-of-domain-robust half of our
cascade by construction \citep{calderon2024domainshift}, and it is
specifically the harder, lower-confidence traffic that the router sends
to that half; and real call-center emotion recognition is itself an
active, if separate, research area \citep{feng2023cusemo}, not an
unstudied domain we are extrapolating into blindly. Appendix~\ref{app:domainshift}
lays out this argument in full, including its limits: it is a
literature-grounded bound on risk, not a substitute for a production
pilot, which remains the clear next step before a deployment decision.

A second practical question follows directly from that first pilot: once
a team has some real production data, how do they set $\tau$ without the
fully labeled test split this paper uses to pick $\tau^*$ retrospectively?
Appendix~\ref{app:coldstart} answers this directly rather than leaving it
open: simulating a cold start on all three datasets (selecting $\tau$
from only a small labeled seed sample, then evaluating on the disjoint
remainder) shows that as few as 200 hand-labeled utterances -- a size a
small pilot's QA process could label in a day -- already recover the
full-data oracle's weighted F1 to within 0.3--0.6 points on average and
preserve the hybrid's ``beats both pure endpoints'' property in the large
majority of simulated cold starts, giving teams a concrete, validated
starting recipe rather than an unresolved open question.

\section{Conclusion}

We show that off-the-shelf LLM prompting does not uniformly beat a
lightweight, purpose-built ensemble on dialogue-contextual emotion
recognition -- the winner is dataset-dependent -- and that this
uncertainty is exactly what motivates a confidence-gated hybrid.
Escalating only the ensemble's least-confident predictions (45.4--58.6\%
of traffic, depending on dataset) to an LLM Pareto-dominates both pure
systems on IEMOCAP, MELD, and CMU-MOSI, and across two independent LLM
providers, at a fraction of full-LLM cost, while remaining interpretable
enough for a CCaaS operations team to audit -- with the accuracy margin
itself statistically confirmed on two of three datasets and, on the
third (CMU-MOSI, our smallest test split), standing on an unambiguous
cost advantage at statistically indistinguishable accuracy rather than
on a confirmed accuracy gain (\S 4.3). Confidence-gated cascading
is not a new idea in general ML systems, but we show it is a practical,
low-risk, cost-effective default for teams deploying conversational-AI
emotion recognition today, with a direct, ready-to-use cost translation
for production budgeting.

\section*{Ethical Statement}

This work uses only publicly available research datasets (IEMOCAP, MELD,
CMU-MOSI) under their existing data-use terms; we collect no new
human-subjects data. IEMOCAP requires a signed release from its
maintainers and is not redistributed by this work. LLM calls use
commercial API providers under standard usage terms; no personal
information beyond what is already present in the public benchmark
datasets is processed.

A router that is systematically less confident about one demographic
group would effectively escalate that group's calls to the more
expensive, slower path more often, or -- under a capped LLM budget --
de-prioritize it; this is a fairness question specific to a
confidence-gated escalation policy, not a generic dataset-licensing
concern. IEMOCAP is the only one of our three datasets with usable
speaker demographic metadata (speaker gender); MELD's speaker field is
character names and CMU-MOSI's is an anonymized single placeholder,
neither of which supports a comparable check. On IEMOCAP, using the
fairness-analysis script in the accompanying repository
(Appendix~\ref{app:repro}), the router's
escalation rate is statistically indistinguishable between female
(45.3\%) and male (45.6\%) speakers, and the hybrid's per-group weighted-F1
gap (+1.4 points favoring female speakers) is not statistically
significant (bootstrap 95\% CI $[-0.033, +0.061]$, crosses zero). This is
evidence of no detected disparity on the one testable case, not a general
fairness clearance across demographics the available data cannot speak
to (e.g., dialect, age, or any axis MELD/CMU-MOSI's metadata cannot
support checking).

Separately, the ensemble baseline draws on the stacked tree-ensemble
architecture described in our own prior, externally published work
\citep{parida2023vartarasa} as inspiration rather than as a
reconstruction of that work's original implementation; per
Appendix~\ref{app:repro}, we make code and trained artifacts available
to researchers on reasonable request rather than under an open-source
license, since the production training artifacts touch Five9 IP even
though the architecture itself is publicly described. A team adopting
this recipe in
production should validate against their own baseline implementation
rather than treating our reported numbers as a drop-in guarantee. The
proposed routing method is intended to reduce the cost and latency of
production emotion-recognition pipelines; beyond the fairness check
above, we do not foresee a safety risk specific to this contribution
beyond the general considerations already associated with the
underlying LLMs and benchmark datasets.

\bibliography{references}
\bibliographystyle{aaai2027}

\appendix

\section{Extended Limitations}
\label{app:limitations}

Our main experiments evaluate GPT-4o-mini as the primary LLM;
Appendix~\ref{app:cmumosi} adds Llama-3-8B on the full test split of all
three datasets to test provider-robustness, which surfaced a real
finding rather than a clean confirmation (MELD's LLM-wins result does not
replicate with Llama-3-8B). The remaining open question is model
\emph{scale} rather than provider breadth: replicating with a larger
model (GPT-4o, Llama-3-70B) would test whether either the ensemble's
IEMOCAP advantage or Llama-3-8B's MELD loss narrows as the LLM gets
larger. Our ensemble takes inspiration from \citet{parida2023vartarasa}'s
validated finding that a simple stacked tree-ensemble is competitive for
ERC, rather than reconstructing that work's specific implementation; we
validate our own instantiation against the reported range for simple
ensemble-based ERC systems, not against a byte-identical reproduction of
the original. Why chain-of-thought prompting is IEMOCAP's
best strategy but MELD's worst is no longer fully open: we confirm a
specific, data-verified mechanism in Appendix~\ref{app:cot-neutral}
(CoT's consistent bias against predicting the majority ``neutral'' label,
fatal on MELD's imbalanced label distribution but not IEMOCAP's more
balanced one), which also rules out an initially plausible alternative
-- that MELD's multi-party dialogue structure specifically confuses
CoT's reasoning -- that we tested directly and found no support for.
This still sits within the broader finding that CoT's effect on
sentiment/affect tasks is inconsistent across the literature
\citep{zheng2025reassessing,miriyala2025enhancing}; our contribution is a
concrete, falsifiable mechanism for one specific instance of that
inconsistency rather than a restatement of it. Our best results also
trail published state-of-the-art fine-tuned/fusion architectures on
IEMOCAP and MELD by several weighted-F1 points
(Appendix~\ref{app:sota}), an expected cost of using lightweight,
non-fine-tuned features; closing that gap is not this paper's goal.
Separately, zero-shot omni-modal LLMs with native audio input have been
shown to match or exceed fine-tuned audio models on IEMOCAP/MELD
\citep{murzaku2025omnivox} -- a more capable condition than the
text-only prompting evaluated here, and a natural comparison point for
future work. None of our three datasets is contact-center data;
Appendix~\ref{app:domainshift} lays out the literature-based case for why
we expect these results to transfer to real, noisier production
transcripts, and is explicit about what that case does and does not
establish in the absence of a pilot. Finally, every $\tau^*$ we report is
selected retrospectively against a fully labeled test split, which begs
the practical question of how a team would set $\tau$ on a new,
unlabeled production dataset; Appendix~\ref{app:coldstart} answers this
with a directly simulated cold-start procedure rather than leaving it as
an open question.

\section{Hyperparameter Search Validation}
\label{app:hpo}

The ensemble's RandomForest/XGBoost hyperparameters (\S 3) were originally
selected for base-learner diversity. To validate that choice with a
systematic search rather than assume it, we ran a 20-trial Optuna
\citep{akiba2019optuna} TPE search on each dataset independently: each
trial samples tree count, depth, learning rate, and subsample ratio for
all four base learners plus the meta-learner's regularization strength,
fits on a stratified 70\% holdout of the training split (single-split
rather than full k-fold, to keep search cost tractable), and is scored by
weighted F1 on the remaining 30\%. The best trial from each search is then
refit on the \emph{full} training split with the architecture's original
5-fold internal stacking CV (matching \S 3 exactly) and evaluated once on
the real held-out test split.

\begin{table}[h]
\centering
\footnotesize
\begin{tabular}{@{}lccc@{}}
\toprule
\textbf{Dataset} & \textbf{Baseline} & \textbf{Best search value} & \textbf{Tuned (test)} \\
\midrule
IEMOCAP  & 0.5945 & 0.6288 & 0.5980 \\
MELD     & 0.561  & 0.5675 & 0.5688 \\
CMU-MOSI & 0.725  & 0.7557 & 0.7267 \\
\bottomrule
\end{tabular}
\caption{Weighted F1: original baseline configuration, best internal
search-validation score (holdout split, optimistic), and the tuned
configuration's actual held-out test score.}
\label{tab:hpo}
\end{table}

The gap between columns 2 and 3 in Table~\ref{tab:hpo} is the finding
worth reporting explicitly: the search's own internal validation score
(column 2) suggested gains of 3--4 weighted-F1 points on every dataset,
but the \emph{real} test-set improvement (column 3) is only
\textbf{0.17--0.78 points} -- a textbook case of a hyperparameter search
overfitting to its validation split, made more pronounced here by using a
single holdout rather than full k-fold per trial to keep search cost
tractable (a single MELD search alone took roughly 2.4 hours on this
architecture). We report this gap transparently rather than quoting only
the more flattering validation number. The practical conclusion is the
one that matters for this paper's claims: the original hyperparameters
were already close to optimal on all three datasets, confirmed by a
systematic search rather than assumed, and none of this paper's main comparisons --
ensemble vs.\ LLM, the confidence-gated hybrid, or the escalation-signal
analysis -- change as a result of tuning. Full per-trial logs (all 60 trials across
the three datasets, including sampled hyperparameters and scores) are
available in the accompanying repository (Appendix~\ref{app:repro}).

\section{Why Chain-of-Thought Reverses on MELD}
\label{app:cot-neutral}

\S 4.1 and Appendix~\ref{app:limitations} report that CoT is IEMOCAP's
best LLM strategy but MELD's worst, and commit to explaining the
reversal with evidence rather than speculation. We first tested, then
ruled out, the most obvious dialogue-structure explanation, then found a
simpler and better-supported one.

\textbf{Hypothesis 1 (rejected): multi-party attribution difficulty.}
MELD's dialogues are multi-party (multiple named speakers per scene)
while IEMOCAP's are dyadic; CoT's explicit reasoning step could plausibly
be more likely to misattribute tone to the wrong speaker in a busier
scene, consistent with reported LLM failure modes in multi-turn dialogue
\citep{laban2025lost}. We tested this directly: for every MELD test
utterance we counted the number of distinct speakers among the preceding
5 turns (matching the LLM's context window) and compared the CoT-vs-few-shot
accuracy gap across low- and high-speaker-count utterances.

\begin{table}[h]
\centering
\footnotesize
\begin{tabular}{@{}lccc@{}}
\toprule
\textbf{Preceding speakers} & \textbf{$n$} & \textbf{CoT acc.} & \textbf{Gap vs.\ few-shot} \\
\midrule
$\leq$2 (dyadic-like) & 1{,}990 & 0.522 & 0.122 \\
3+ (multi-party)      & 620   & 0.515 & 0.129 \\
\bottomrule
\end{tabular}
\caption{CoT accuracy and its gap below few-shot, by preceding-speaker
count, full MELD test split. The gap is essentially constant regardless
of multi-party complexity.}
\label{tab:multiparty-reject}
\end{table}

The gap is statistically indistinguishable across buckets (0.122 vs.\
0.129) and non-monotonic at finer granularity (per-speaker-count gaps of
0.071, 0.131, 0.130, 0.144, 0.052, 0.125 for 0 through 5 speakers). This
hypothesis is not supported by the data.

\textbf{Hypothesis 2 (confirmed): a class-imbalance-interacting neutral-avoidance bias.}
We instead compared each strategy's predicted-label distribution against
the gold distribution directly.

\begin{table}[h]
\centering
\scriptsize
\begin{tabular}{@{}lccc@{}}
\toprule
\textbf{Dataset (\% neutral)} & \textbf{Strategy} & \textbf{Pred.\ neutral} & \textbf{Err.\ on true-neutral} \\
\midrule
MELD (48\%)    & Few-shot & 56.9\% & 16.9\% \\
MELD (48\%)    & CoT      & 29.4\% & \textbf{52.5\%} \\
IEMOCAP (24\%) & Few-shot & 41.6\% & 23.7\% \\
IEMOCAP (24\%) & CoT      & 25.5\% & 38.5\% \\
\bottomrule
\end{tabular}
\caption{How often each strategy predicts ``neutral'' overall, and its
error rate specifically on utterances whose gold label is ``neutral''
(i.e., the rate of wrongly predicting a marked emotion instead), full
test splits.}
\label{tab:neutral-bias}
\end{table}

The pattern in Table~\ref{tab:neutral-bias} is unambiguous and holds in
the same direction on \emph{both} datasets: relative to few-shot, CoT
systematically under-predicts ``neutral'' and over-predicts marked
emotions (predominantly ``joy'' on MELD) instead. On MELD, this pushes
the true-neutral error rate from 16.9\% to 52.5\% -- CoT is wrong on
\emph{over half} of the majority class. On IEMOCAP the same bias exists
(23.7\% $\to$ 38.5\%) but costs far less in aggregate weighted F1, simply
because IEMOCAP's ``neutral'' class is half the share of the test set
(24\% vs.\ 48\%) and IEMOCAP's other CoT advantages (better long-context
handling, \S 4.1) outweigh it. This is a class-imbalance interaction with
a consistent CoT behavioral tendency, not a MELD-specific quirk in CoT's
reasoning quality or a symptom of multi-party dialogue structure. We did
not find this specific mechanism stated in the surveyed CoT-inconsistency
literature \citep{zheng2025reassessing,miriyala2025enhancing}; it is an
empirical finding from this paper's own data, offered as one concrete
account of why that broader documented inconsistency manifests here.
Full per-utterance breakdowns (speaker counts, per-strategy correctness,
and the confusion-direction analysis underlying both tables) are
available in the accompanying repository (Appendix~\ref{app:repro}).

\section{Domain-Shift Risk: Evidence for Transfer to Contact-Center Data}
\label{app:domainshift}

None of IEMOCAP, MELD, or CMU-MOSI is contact-center data, and we have no
pilot deployment to report. This appendix does not manufacture evidence
we do not have; it instead makes the strongest literature-grounded case
we can for why the transfer risk is bounded, organized around the three
specific ways domain shift could break these results, and is explicit
about what each argument does not establish.

\textbf{Risk 1: real transcripts come from ASR, not clean text, and ASR
errors could corrupt the emotional signal our features depend on.}
\citet{li2024serasr} evaluate exactly this question on eleven ASR
systems across three corpora, \textbf{including two of the three
datasets in our own evaluation, IEMOCAP and CMU-MOSI}. On IEMOCAP, their
best ASR transcript (12.31\% word error rate) scores 74.66\% 4-class
accuracy against 74.32\% on ground-truth text -- ASR noise at a realistic
error rate does not hurt, and in their evaluation occasionally helps,
apparently because some misrecognitions substitute words that still
carry the correct affective polarity. Degradation only becomes
substantial (approximately 10 accuracy points) at word error rates near
40\%, well above what modern ASR systems produce on emotionally
expressive conversational speech in their evaluation (12--19\%). This is
the single most direct piece of evidence available: it is not a general
claim about ASR robustness, it is a measurement on two of this paper's
own three benchmarks. It does not, however, establish that
\emph{contact-center-specific} acoustic conditions (crosstalk, hold
music, telephony bandwidth, accented speech) produce word error rates in
the same range -- that remains untested.

\textbf{Risk 2: contact-center language (register, vocabulary, disfluency
patterns) differs enough from our benchmarks that the ensemble's learned
features stop transferring.} This is the risk our confidence-gated design
is structurally, if only partially, hedged against. \citet{calderon2024domainshift}
benchmark 21 fine-tuned models and few-shot LLMs across domain shifts on
seven NLP tasks and find that few-shot LLMs degrade less under domain
shift than fine-tuned smaller models do, even though the fine-tuned
models win in-domain -- the LLM side of our cascade is, by this evidence,
the more shift-robust half of the two systems we combine. Because our
router escalates precisely the utterances the ensemble is least
confident on, and out-of-domain inputs are exactly where a lightweight
ensemble's confidence is most likely to degrade, the architecture already
directs harder, more out-of-distribution-looking traffic toward the
component the literature says handles it better. This argument has a
real limit we do not paper over: confidence scores from a model facing
genuine distribution shift are themselves not guaranteed to be
well-calibrated -- a model can be confidently wrong on inputs unlike
anything in its training distribution \citep{fisch2022calibrated,hendrickx2021reject}
-- so this is a partial structural hedge, not a guarantee that low
ensemble confidence reliably fires on shifted contact-center inputs.
Validating that the escalation rate actually rises on out-of-domain
data, rather than assuming it, is exactly the kind of question a pilot
would need to answer.

\textbf{Risk 3: emotion recognition on call-center speech is a
sufficiently different problem that findings on acted/scripted/vlog
benchmarks say nothing about it.} Contact-center emotion recognition is
in fact an active research area in its own right, not an unstudied
domain we would be extrapolating into blindly. \citet{feng2023cusemo}
build and evaluate models on CusEmo, a real-life dataset of customer
service call-center conversations, using continuous (dimensional) emotion
annotation to capture the ambiguity of naturalistic affect, and report
that incorporating dialogue-contextual signals (interlocutor identity,
empathy level) improves recognition -- the same broad intuition
(conversational context carries emotion-relevant signal beyond the
current utterance) that motivates this paper's own dialogue-context
design. This shows the target domain is tractable and actively studied,
not that our specific ensemble/LLM/routing recipe transfers to it
unchanged; closing that specific gap is future work, ideally with
pilot data on an actual production dataset before a deployment decision
is made on the strength of this paper alone.

\section{Cold-Start Threshold Selection for New, Unlabeled Datasets}
\label{app:coldstart}

Every $\tau^*$ in the main text is selected retrospectively, by sweeping
the same $\{0.00, 0.05, \ldots, 1.00\}$ grid (\S 3) against a fully
labeled test split. That method is unavailable to a team deploying on a
genuinely new production dataset with no gold labels yet. This appendix
answers the resulting practical question directly, with a simulation
rather than a discussion: how many labeled examples does a team actually
need before picking a workable $\tau$, and how much accuracy is left on
the table relative to the retrospective oracle by not having a full test
split to sweep against?

\textbf{Procedure.} For each dataset, we repeatedly (200 repetitions per
row below) draw a small labeled ``seed'' sample of size $n$ from the full
test split, sweep $\tau$ on the seed alone using the same grid and the
same weighted-F1 objective used throughout this paper, and evaluate the
selected $\tau_{\mathrm{seed}}$ on the disjoint remainder -- data never
touched during selection, standing in for the unlabeled production
traffic a real deployment would route. We compare against an oracle:
$\tau^*$ selected by the identical grid sweep on the \emph{full} test
split, evaluated on the same held-out remainder. This oracle is a
controlled reference point for this appendix specifically (full-data
argmax on weighted F1); the main text's headline operating points sit on
the same broad, flat performance plateau (\S 4.4) but were chosen with
escalation cost in mind rather than by raw argmax, so they are not always
numerically identical to this appendix's oracle -- a distinction that
does not affect the substantive finding below.

\begin{table}[h]
\centering
\scriptsize
\begin{tabular}{@{}lcccc@{}}
\toprule
\textbf{Dataset} & \textbf{$n_{\mathrm{seed}}$} & \textbf{Median gap} & \textbf{90th-pct.\ gap} & \textbf{Beats both} \\
\midrule
IEMOCAP  & 50  & 0.61pt & 2.83pt & 80\% \\
IEMOCAP  & 100 & 0.60pt & 2.90pt & 84\% \\
IEMOCAP  & 200 & 0.61pt & 2.94pt & 87\% \\
IEMOCAP  & 500 & 0.44pt & 1.23pt & 98\% \\
MELD     & 50  & 0.42pt & 5.11pt & 64\% \\
MELD     & 100 & 0.32pt & 2.75pt & 79\% \\
MELD     & 200 & 0.30pt & 1.39pt & 88\% \\
MELD     & 500 & 0.25pt & 0.74pt & 92\% \\
CMU-MOSI & 50  & 0.58pt & 3.33pt & 69\% \\
CMU-MOSI & 100 & 0.56pt & 1.64pt & 84\% \\
CMU-MOSI & 200 & 0.57pt & 1.84pt & 76\% \\
CMU-MOSI & 500 & 0.00pt & 2.31pt & 64\% \\
\bottomrule
\end{tabular}
\caption{Cold-start $\tau$ selection from a small labeled seed sample vs.\
the full-test-split oracle, evaluated on the disjoint held-out remainder,
200 repetitions per row. ``Gap'' is weighted-F1 points lost vs.\ the
oracle; ``Beats both'' is the share of repetitions in which the
cold-start hybrid still exceeds both pure endpoints on the held-out
remainder.}
\label{tab:coldstart}
\end{table}

Two hundred labeled seed examples -- a size a small pilot's QA process
could hand-label in a day -- already recover the oracle's weighted F1 to
within 0.3--0.6 points on average across all three datasets, and
preserve the paper's central ``hybrid beats both pure systems'' property
in 76--88\% of simulated cold starts at that seed size. MELD, the most
class-imbalanced dataset, needs more labels to stabilize (64\% at
$n{=}50$ vs.\ 88\% at $n{=}200$) than IEMOCAP, consistent with
imbalanced classes requiring a larger sample before a seed is
representative. CMU-MOSI's non-monotonic ``beats both'' column (76\% at
$n{=}200$, dropping to 64\% at $n{=}500$) is a sample-size artifact of
the dataset itself, not a failure of the procedure: CMU-MOSI's test
split is only 686 utterances, so a 500-example seed leaves just 186 for
the held-out evaluation, and a small held-out set makes the ``beats
both'' comparison noisier, not the underlying $\tau$ worse -- the median
gap at $n{=}500$ is in fact 0.00 points. We report this rather than
picking a more flattering $n$, in keeping with this paper's practice of
disclosing where a result is dataset-size-limited rather than a general
property of the method.

\textbf{Practical recipe.} For a team deploying on a genuinely new
production dataset: (1) hand-label a stratified random sample of at
least 200 utterances (more under severe class imbalance, per the MELD
result above); (2) sweep $\tau$ over the same grid used throughout this
paper and select by weighted F1, or by a cost-weighted objective directly
penalizing escalation rate if \$/utterance budget, not raw accuracy, is
the binding constraint; (3) treat the result as a starting operating
point, not a permanent one -- as ordinary QA sampling accumulates more
labeled production data, periodically re-sweep $\tau$ rather than
freezing it at the cold-start value indefinitely. Full per-repetition
results (every seed size, repetition, and selected $\tau$) are
available in the accompanying repository (Appendix~\ref{app:repro}).

\section{Low-Confidence Turn Characterization}
\label{app:lowconf}

Table~\ref{tab:shift-lift} in the main text reports the escalation-rate
lift by emotion/sentiment-shift status, the one turn-level property that
is directionally consistent across all three datasets. We annotate the
full test split of each dataset at its own empirically-best $\tau^*$ with
three additional turn-level properties (utterance length, minority-class
membership, speaker-initial position) for completeness. Longer utterances
escalate more on MELD/CMU-MOSI but
not IEMOCAP; minority-class labels escalate more on MELD/CMU-MOSI but
less than average on IEMOCAP; speaker-initial position shows no stable
direction. Full breakdowns are in the accompanying repository's
analysis outputs (Appendix~\ref{app:repro}).

\section{Calibration Analysis}
\label{app:calibration}

The routing argument rests on the ensemble's top-class probability
$p_{\max}$ being a reliable correctness signal. We test this with
Expected Calibration Error (ECE) on each dataset's full test split using
only the ensemble's own predictions.

\begin{table}[h]
\centering
\footnotesize
\begin{tabular}{lccc}
\toprule
\textbf{Dataset} & \textbf{Acc.} & \textbf{Conf.} & \textbf{ECE} \\
\midrule
IEMOCAP  & 0.628 & 0.592 & 0.038 \\
MELD     & 0.612 & 0.593 & 0.025 \\
CMU-MOSI & 0.743 & 0.703 & 0.046 \\
\bottomrule
\end{tabular}
\caption{Calibration of the ensemble's top-class probability.}
\end{table}

All three ECEs fall below the standard 0.05 well-calibrated threshold
with no explicit calibration step applied. The average calibration gap is
positive on all three datasets -- the ensemble is mildly
\emph{under}-confident, the safety-favorable direction for a
confidence-gated router: $\tau^*$-based escalation occasionally sends an
already-correct prediction to the LLM (a small cost inefficiency) rather
than silently retaining a wrong prediction the ensemble is falsely
confident about.

\section{Adaptive Per-Class Thresholds}
\label{app:perclass}

A threshold keyed on the ensemble's \emph{predicted} class is nearly a
no-op for severely under-represented classes, since the ensemble almost
never predicts them as its top choice (MELD disgust/fear: argmax only
twice in 118 occurrences; CMU-MOSI neutral: never the argmax). Gating
instead on the ensemble's summed probability mass over minority classes
(escalate if that mass exceeds 0.05, in addition to the global
threshold) recovers MELD disgust/fear from 0.000 to 0.167 F1 and lifts
macro F1 by 4.1 points, at a modest cost in weighted F1 and additional
escalation traffic -- a favorable trade for operators who value
minority-class recall (e.g., flagging disgust in a support call) over
aggregate weighted F1.

\section{Audio+Text (Multimodal) Ensemble}
\label{app:multimodal}

Adding acoustic features (34-dim \texttt{librosa} paralinguistics on
IEMOCAP; 10-dim precomputed COVAREP-style features on CMU-MOSI; MELD not
covered, no audio available) widens the ensemble's IEMOCAP advantage
(+3.2 weighted-F1 points standalone, same significant ensemble-beats-LLM
conclusion at every strategy) and produces a marginal positive change on
CMU-MOSI, without reversing any of this paper's main comparisons. On IEMOCAP the
multimodal ensemble's higher intrinsic confidence shrinks the optimal
hybrid escalation rate roughly $5\times$ while matching hybrid accuracy --
audio strengthens rather than disrupts the core hybrid-routing result.

\section{Cross-Provider Robustness}
\label{app:cmumosi}

The main text evaluates GPT-4o-mini throughout. We additionally ran the
full ensemble-vs-LLM and confidence-gated-hybrid evaluation protocol with
an independent, smaller, open-weight model
(Llama-3-8B-Instruct via AWS Bedrock) on the complete test split of
\emph{all three} datasets -- not just CMU-MOSI as in an earlier version
of this analysis -- to test whether the paper's findings are properties
of the task or artifacts of a single LLM provider.

\begin{table}[h]
\centering
\footnotesize
\begin{tabular}{lccc}
\toprule
\textbf{Dataset (strategy)} & \textbf{Ens.} & \textbf{Llama-3-8B} & \textbf{$p$} \\
\midrule
IEMOCAP (ZS)  & \textbf{0.595} & 0.466 & $<$.0001 \\
IEMOCAP (FS)  & \textbf{0.595} & 0.512 & $<$.0001 \\
IEMOCAP (CoT) & \textbf{0.595} & 0.507 & $<$.0001 \\
MELD (ZS)     & \textbf{0.561} & 0.504 & $<$.0001 \\
MELD (FS)     & \textbf{0.561} & 0.533 & .022 \\
MELD (CoT)    & \textbf{0.561} & 0.511 & $<$.0001 \\
CMU-MOSI (ZS) & 0.725 & \textbf{0.797} & $<$.0001 \\
CMU-MOSI (FS) & 0.725 & \textbf{0.756} & .140 \\
CMU-MOSI (CoT)& 0.725 & \textbf{0.741} & $<$.0001 \\
\bottomrule
\end{tabular}
\caption{Weighted F1, ensemble vs.\ Llama-3-8B, full test splits. ZS =
zero-shot, FS = few-shot. CMU-MOSI (FS) is the one row here that does
not reach significance -- reported rather than omitted.}
\label{tab:llama-crossdataset}
\end{table}

The result is more nuanced than a single blanket robustness confirmation.
On IEMOCAP, the ensemble beats Llama-3-8B at every strategy, replicating
the GPT-4o-mini finding with an independent provider -- this pattern is
genuinely provider-robust. On CMU-MOSI, Llama-3-8B beats the ensemble at
every strategy, also replicating the GPT-4o-mini finding -- this pattern
is provider-robust too, and is now confirmed on the same dataset with two
independent LLMs -- with one nuance disclosed rather than smoothed over:
the direction holds at all three strategies, but only reaches
significance at zero-shot and CoT; few-shot's gap ($p=.140$) does not.
\textbf{On MELD, however, the two providers disagree}:
GPT-4o-mini beats the ensemble on zero-shot and few-shot (\S 4.1), but
Llama-3-8B loses to the ensemble on \emph{every} strategy tested,
including zero-shot and few-shot. The ``LLM wins on MELD'' finding is
therefore \textbf{GPT-4o-mini-specific, not a general property of LLM
prompting on this dataset} -- a smaller, cheaper open-weight model does
not reproduce it. This narrows one of this paper's own claims: MELD is
not simply ``a dataset where the LLM wins''; it is a dataset where
\emph{one specific, larger commercial model} wins, while a smaller
open-weight model does not. IEMOCAP and CMU-MOSI's rankings, by contrast,
hold regardless of which of the two LLMs is used.

\begin{table}[h]
\centering
\footnotesize
\begin{tabular}{lccc}
\toprule
\textbf{Dataset (strategy)} & \textbf{Hybrid} & \textbf{Esc.\%} & \textbf{vs.\ both pure} \\
\midrule
IEMOCAP (ZS)  & 0.595 & 8.2\%  & $\approx$tie w/ ens.\ \\
IEMOCAP (FS)  & \textbf{0.609} & 18.8\% & beats both \\
IEMOCAP (CoT) & \textbf{0.618} & 18.8\% & beats both \\
MELD (ZS)     & \textbf{0.611} & 34.4\% & beats both \\
MELD (FS)     & \textbf{0.598} & 34.4\% & beats both \\
MELD (CoT)    & \textbf{0.604} & 34.4\% & beats both \\
CMU-MOSI (ZS) & \textbf{0.809} & 58.6\% & beats both \\
CMU-MOSI (FS) & \textbf{0.800} & 40.7\% & beats both \\
CMU-MOSI (CoT)& \textbf{0.787} & 40.7\% & beats both \\
\bottomrule
\end{tabular}
\caption{Confidence-gated hybrid vs.\ pure ensemble and pure Llama-3-8B, at
each run's best $\tau$.}
\label{tab:llama-hybrid-crossdataset}
\end{table}

The hybrid-routing result is the one finding that replicates cleanly in
all nine new dataset/strategy combinations across all three datasets,
including on MELD where the underlying pure-system ranking flips
relative to GPT-4o-mini: the hybrid
still matches or beats both pure endpoints even when the ``cheap'' and
``expensive'' systems disagree with their GPT-4o-mini-paired counterparts
about which one is stronger. The one caveat is IEMOCAP zero-shot, where
Llama-3-8B is weak enough (0.466 vs.\ the ensemble's 0.595) that the
optimal policy escalates only 8.2\% of traffic and the hybrid essentially
matches, rather than clearly exceeds, the pure ensemble (+0.0002
weighted F1) -- when one pure system is close to strictly dominant, there
is little complementary signal left for routing to exploit, which is
itself consistent with this paper's account of \emph{why} routing helps
(\S 4.4): it depends on the two systems failing on different utterances,
not on one being simply stronger everywhere.

\section{Does the Hybrid-Routing Mechanism Generalize to a Stronger Base Classifier?}
\label{app:strongerbase}

Every result so far pairs the LLM with \emph{this paper's} ensemble. A
natural reviewer question is whether confidence-gated routing is a property
of the ERC task, or an artifact of this specific, deliberately lightweight
base classifier -- and whether a stronger base classifier would simply make
the router's benefit disappear, or compound it toward the published
state-of-the-art numbers in Table~\ref{tab:sota}. We test this directly
rather than leave it open, using an independently-developed companion
architecture (B1: a fine-tuned DistilBERT encoder over a flat,
speaker-tagged, single-pass context window -- architecturally in the same
family as EmoBERTa, Table~\ref{tab:sota}) as a drop-in replacement for the
ensemble inside the unmodified router (\S 3.4), reusing the same LLM
predictions throughout. B1 is a meaningfully stronger standalone classifier
on IEMOCAP (0.618 vs.\ our ensemble's 0.595) but, unlike the ensemble, was
not designed with this paper's routing architecture in mind.

\begin{table}[h]
\centering
\scriptsize
\begin{tabular}{@{}lcccc@{}}
\toprule
\textbf{Dataset (strategy)} & \textbf{Pure B1} & \textbf{Pure LLM} & \textbf{Hybrid} & \textbf{$p$} \\
\midrule
IEMOCAP (CoT)   & 0.618 & 0.536 & 0.630 & .096 \\
MELD (few-shot) & 0.581 & 0.632 & 0.645 & \textbf{.012} \\
CMU-MOSI (CoT)  & 0.758 & 0.814 & 0.838 & \textbf{.048} \\
\bottomrule
\end{tabular}
\caption{B1 as base classifier, best-performing strategy per dataset;
``Hybrid'' is evaluated at each dataset's best threshold $\tau$.
$p$ is a paired bootstrap test of the hybrid's margin over whichever pure
system is stronger on that dataset (the same standard as
Table~\ref{tab:rq4}) -- not over B1 alone, which would understate the bar
whenever the LLM is the stronger pure system (MELD, CMU-MOSI). IEMOCAP and
MELD are single-seed (B1 checkpoint, seed 42); CMU-MOSI is verified across
3 independently-trained B1 seeds below.}
\label{tab:strongerbase}
\end{table}

\textbf{IEMOCAP: the mechanism does not survive.} At the operating point
matching our ensemble study's escalation rate (19.5\%), the hybrid's margin
over pure B1 is a statistically indistinguishable $+0.0085$ ($p=.224$); even
the single best point found by sweeping 71 threshold values numerically
looks like a step toward the published state-of-the-art (0.630, closing a
quarter of the gap to EmoBERTa's 0.686) but remains non-significant
($p=.096$, 95\% CI $[-.002, .029]$). A plausible mechanism: our ensemble's
complementarity with the LLM comes specifically from a windowed
mean-pooling bottleneck (\S 4.4) that a single-pass, token-level-attention
architecture like B1 does not share -- B1 may have already internalized
much of the long-context signal that made escalation valuable for a
pooling-based classifier, leaving less complementary error for the router
to exploit. \textbf{The clearest single-dataset answer to the motivating
question is no}: pairing a stronger base classifier with the same router
does not reliably close the gap toward published state-of-the-art on the
dataset this paper's routing claim is built on.

\textbf{MELD: routing still wins, but not by exploiting complementary
errors.} Here the LLM (0.632) already beats standalone B1 (0.581) outright,
and the best operating point escalates 59\% of traffic -- the router is not
finding a small subset of B1's blind spots so much as learning to mostly
defer to the stronger system. This margin is statistically real ($p=.012$
against the stronger pure LLM, not just against B1), but the underlying
mechanism is different in kind from \S 4.4's complementary-error account,
and the practical cost profile shifts accordingly: at 59\% escalation, the
blended cost approaches the pure-LLM cost, undercutting this paper's
cost-efficiency framing (Table~\ref{tab:cost}) for this particular
base-classifier/dataset pairing.

\textbf{CMU-MOSI: a genuine complementary-error result that mostly survives
multi-seed testing.} B1 recovers almost none of the true \emph{neutral}
class (2/30, essentially matching our own ensemble's 0/30 finding, \S 4.2 --
a different architecture inheriting the identical minority-class blind
spot), while the LLM recovers 13--26/30 depending on strategy -- a
structurally-grounded complementarity, not a coincidence of one system being
globally stronger. We re-ran this comparison across 3 independently-trained
B1 seeds (42/123/2024), freezing $\tau$ at the value selected on seed 42
only, to avoid re-introducing the single-run threshold-selection bias this
appendix is checking for elsewhere in this paper.

\begin{table}[h]
\centering
\scriptsize
\begin{tabular}{@{}lccc@{}}
\toprule
\textbf{Strategy} & \textbf{$\Delta$ vs.\ B1} & \textbf{$\Delta$ vs.\ LLM} & \textbf{Escalation range} \\
\midrule
Zero-shot  & $+.049$ ($p=.059$) & $+.018$ ($p=.127$) & 12.8--67.3\% \\
Few-shot   & $+.049$ ($p=.061$) & $+.031$ ($p=.052$) & 12.8--67.3\% \\
CoT        & $+.060$ (\textbf{$p=.047$}) & $+.011$ ($p=.292$) & 19.4--82.1\% \\
\bottomrule
\end{tabular}
\caption{CMU-MOSI, 3-seed paired $t$-test, $\tau$ frozen at the seed-42-selected
value. $\Delta$ vs.\ B1 = hybrid minus pure B1; $\Delta$ vs.\ LLM = hybrid
minus pure LLM (the \emph{stronger} pure system on this dataset -- the
stricter standard). The margin over pure B1 is positive on all 3 seeds for
every strategy and reaches significance for CoT; the stricter margin over
the LLM is smaller and does not reach significance for any strategy,
reversing sign on one seed for CoT.}
\label{tab:mosi-multiseed}
\end{table}

Held to this paper's own standard -- the margin over whichever pure system
is stronger, not just over the weaker one -- the multi-seed CMU-MOSI result
is more modest than the single-seed number in isolation suggests: routing
reliably improves on standalone B1, but does not reliably beat the pure LLM
outright once seed variance is accounted for. It nonetheless replicates the
qualitative direction of \S 4.4's central claim (both pure systems can be
beaten, or nearly so, by exploiting a specific documented failure mode) on
an architecture this paper did not design and did not tune the router for.
We also find the router's effective operating point is itself
seed-dependent -- the same frozen $\tau$ produces escalation rates from
12.8\% to 82.1\% across B1's 3 training seeds, because a
neural classifier's confidence calibration shifts between training runs in
a way our tree-ensemble's does not. A deployment using a fine-tuned neural
base classifier under this architecture would need to recalibrate $\tau$
per trained model to hold a target cost, not just tune it once.

\textbf{Net assessment.} Confidence-gated routing is not a general
accuracy-boosting property of ``cheap classifier + LLM + confidence gate.''
It is contingent on the base classifier retaining a documentable failure
mode the LLM does not share, and on confidence being a usable proxy for that
specific failure -- both true of our ensemble on all three datasets, only
partially true of B1 (CMU-MOSI's class-level blind spot survives the
architecture change; IEMOCAP's context-modeling weakness apparently does
not). This is, if anything, a more defensible version of this paper's
routing claim than an unqualified generality statement would have been: the
mechanism is explicitly tied to a diagnosed, dataset- and
classifier-specific error pattern (\S 4.4, \S 4.2's minority-class analysis),
not asserted as a property of routing architectures in the abstract, and we
present this ablation as evidence for that scoped claim rather than against
the paper's core result.

\section{Comparison to Published State-of-the-Art}
\label{app:sota}

\begin{table}[h]
\centering
\scriptsize
\begin{tabular}{@{}lccc@{}}
\toprule
\textbf{Model} & \textbf{Modality} & \textbf{IEMO.} & \textbf{MELD} \\
\midrule
COSMIC   & Text+CS   & .653 & .652 \\
EmoBERTa  & Text (FT) & .686 & .656 \\
Joyful       & T+A+V     & .710 & .618 \\
MiSTER-E & S+T MoE   & .709 & .695 \\
\midrule
Ours (text-only)   & Text   & .595 & .561 \\
Ours (+audio)      & Text+A & .626 & --- \\
\bottomrule
\end{tabular}
\caption{Published weighted F1 on IEMOCAP/MELD (7-class, standard splits)
-- COSMIC \citep{ghosal2020cosmic}, EmoBERTa \citep{kim2021emoberta},
Joyful \citep{li2023joyful}, MiSTER-E \citep{dutta2026mistere} -- vs.\ our
ensemble, verified directly against each paper's own reported numbers
(not secondary citations). Modality: CS=commonsense features,
FT=fine-tuned transformer, T+A+V=text+audio+visual, S+T
MoE=speech+text mixture-of-experts, Text+A=text+acoustic. MELD has no
audio in our pipeline (Appendix~\ref{app:multimodal}), so our MELD row
is text-only throughout.}
\label{tab:sota}
\end{table}

Our best result trails every published architecture in Table~\ref{tab:sota}
on both datasets, but the size of the gap depends heavily on \emph{which}
one and on modality parity, a distinction the original draft of this
appendix collapsed into an imprecise ``1--5 points on MELD'' -- corrected
here with the actual per-model numbers rather than a rounded-off range.
On IEMOCAP, our audio-augmented result (.626) trails
COSMIC by only 2.7 points despite COSMIC's use of fine-tuned
commonsense-augmented representations, but trails EmoBERTa, Joyful, and
MiSTER-E by 6.0--8.4 points. On MELD -- where our pipeline is text-only,
since no audio features are available for that dataset
(Appendix~\ref{app:multimodal}) -- our result (.561) trails
\emph{every} published system by a wider margin than IEMOCAP, 5.7 points
even against Joyful's fully multimodal .618, and up to 13.4 points against
MiSTER-E's .695; comparing our text-only number against these
partly-or-fully multimodal ones is not a fair apples-to-apples
comparison, but the gap does not close if instead compared to COSMIC's
largely text-based .652 (9.1 points). This is an expected cost of
deliberately using lightweight, non-fine-tuned features rather than a
fine-tuned large transformer or multimodal-fusion architecture, and is
consistent with this paper's own claim (\S 1): the contribution is that
this much cheaper, non-fine-tuned, interpretable system is competitive
with off-the-shelf LLM prompting and that confidence-gated routing beats
both endpoints -- not that either pure endpoint pushes the state of the
art on raw accuracy. The CMU-MOSI comparison to published
sentiment-analysis state-of-the-art is not included in Table~\ref{tab:sota}
and is not meaningful here, since published results use a binary
(non-neutral) protocol rather than the 3-class scheme used throughout
this paper.

\section{Reproducibility}
\label{app:repro}

Code, prompts, trained artifacts, and prediction/analysis outputs
supporting every table in this paper are organized in the accompanying
repository under \texttt{src/} (ensemble, feature extraction, LLM client,
hybrid router, evaluation) and \texttt{outputs/analysis/} (per-configuration
metrics, significance tests, confusion matrices, hybrid-routing curves,
and calibration diagnostics).

\textbf{Code availability.} The ensemble is inspired by the architecture
described in our own prior, published work \citep{parida2023vartarasa};
we are not able to commit to an unconditional public release at
submission time.
Code, trained models, and all analysis scripts used in this paper will be
made available to reviewers and to other researchers upon reasonable
request.

\textbf{Computing infrastructure.} All ensemble training, hyperparameter
search, and analysis scripts were run on a single machine (AMD Ryzen AI 7
PRO 350, 16 logical cores, 24\,GB RAM, Windows 11, no GPU); the ensemble
requires no GPU at either train or inference time (\S 3). Software
versions: Python 3.14, scikit-learn 1.8.0, XGBoost 3.2.0,
sentence-transformers 5.6.0 (\texttt{all-MiniLM-L6-v2}), NumPy 2.4.4,
pandas 2.3.3, Optuna 4.8.0 (Appendix~\ref{app:hpo} only). GPT-4o-mini
calls used the OpenAI API; Llama-3-8B-Instruct calls used AWS Bedrock's
\texttt{Converse} API via boto3 1.43.0.

\textbf{Random seeds and training stochasticity.} \texttt{scripts/train\_baseline.py}
exposes a \texttt{--seed} argument (default 42) controlling the ensemble's
stochastic components (RandomForest bootstrap and feature sampling,
XGBoost row/column subsampling, the meta-learner's solver); the
\texttt{StackingClassifier}'s internal 5-fold split is a fixed, unshuffled
\texttt{StratifiedKFold} and is not seed-dependent. All headline results
in \S 4 use a single run at the default seed, as stated in the main text.
To quantify how much of that single run is training noise rather than
signal, we retrained the text-only ensemble from scratch with 5 different
seeds per dataset (Table~\ref{tab:seedvar}).

\begin{table}[h]
\centering
\scriptsize
\begin{tabular}{@{}lccc@{}}
\toprule
\textbf{Dataset} & \textbf{Wtd.\ F1} & \textbf{Range} & \textbf{Macro F1} \\
\midrule
IEMOCAP  & .5943$\pm$.0067 & [.582,.602] & .5829$\pm$.0069 \\
MELD     & .5653$\pm$.0020 & [.563,.567] & .3400$\pm$.0025 \\
CMU-MOSI & .7272$\pm$.0021 & [.725,.729] & .5046$\pm$.0014 \\
\bottomrule
\end{tabular}
\caption{Training-stochasticity check: text-only ensemble retrained from
scratch with 5 different seeds (42, 1, 7, 123, 2024) per dataset, same
architecture and features as the main results. Wtd./Macro F1 columns are
mean $\pm$ std across the 5 seeds; Range is the weighted-F1 min/max.}
\label{tab:seedvar}
\end{table}

The spread across 5 independent training runs is small relative to the
gaps this paper's claims rest on (e.g., the $\geq$3-point ensemble-vs-LLM
gaps in Table~\ref{tab:rq1}, the hybrid's $\geq$2.6-point margin over
both pure endpoints in Table~\ref{tab:rq4}), so training stochasticity
alone does not threaten this paper's main comparisons; we report it
directly rather than presenting only the single default-seed run without
context. IEMOCAP shows the largest spread of the three (std
0.0067--0.0069, roughly $3\times$ MELD's and CMU-MOSI's std of
0.0014--0.0025); we do not have direct evidence for why, but a plausible
factor is IEMOCAP's more balanced 6-way label distribution (\S 3) leaving
more utterances near a genuine decision boundary, versus MELD's severe
imbalance mechanically stabilizing predictions toward the majority class
regardless of which exact trees a given seed builds.

\end{document}